\documentclass[conference]{IEEEtran}

\usepackage[frozencache,cachedir=.]{minted}
\usepackage{amsmath,amsfonts}
\usepackage{algorithmic}
\usepackage{algorithm}
\usepackage{array}
\usepackage[caption=false,font=footnotesize]{subfig}
\usepackage{textcomp}
\usepackage{stfloats}
\usepackage{url}
\usepackage{verbatim}
\usepackage{graphicx}
\usepackage{cite}

\usepackage{newfloat}
\usepackage{listings}
\usepackage{xcolor}
\usepackage{multirow}
\usepackage{multicol}
\usepackage{booktabs}
\usepackage{adjustbox}
\usepackage{ulem}
\usepackage{subcaption}
\usepackage{amsmath}
\usepackage{colortbl}
\usepackage{enumitem}
\usepackage{minted}
\usepackage{pifont}
\usepackage{hyperref}
\usepackage[algo2e,ruled]{algorithm2e} 
\usepackage{makecell}

\IEEEoverridecommandlockouts

\begin{document}

\title{Efficient User Sequence Learning for Online Services via Compressed Graph Neural Networks}

\author{
\IEEEauthorblockN{
Yucheng Wu\IEEEauthorrefmark{1}\IEEEauthorrefmark{2},
Liyue Chen\IEEEauthorrefmark{1}\IEEEauthorrefmark{2},
Yu Cheng\IEEEauthorrefmark{3}, 
Shuai Chen\IEEEauthorrefmark{3},
Jinyu Xu\IEEEauthorrefmark{3},
Leye Wang\IEEEauthorrefmark{1}\IEEEauthorrefmark{2}\thanks{ Leye Wang and Shuai Chen are corresponding authors.}}
\IEEEauthorblockA{\IEEEauthorrefmark{1}Key Lab of High Confidence Software Technologies (Peking University), Ministry of Education, Beijing, China}
\IEEEauthorblockA{\IEEEauthorrefmark{2}School of Computer Science, Peking University, Beijing, China}
\IEEEauthorblockA{\IEEEauthorrefmark{3}Alipay (Hangzhou) Information \& Technology Co. Ltd., Hangzhou, China}
\IEEEauthorblockA{wuyucheng@stu.pku.edu.cn, chenliyue2019@gmail.com, \{cy122623, shuai.cs, daihe.xjy\}@antgroup.com, leyewang@pku.edu.cn}
}

\maketitle

\normalem  

\begin{abstract}
Learning representations of user behavior sequences is crucial for various online services, such as online fraudulent transaction detection mechanisms. Graph Neural Networks (GNNs) have been extensively applied to model sequence relationships, and extract information from similar sequences. While user behavior sequence data volume is usually huge for online applications, directly applying GNN models may lead to substantial computational overhead during both the training and inference stages and make it challenging to meet real-time requirements for online services. In this paper, we leverage graph compression techniques to alleviate the efficiency issue. Specifically, we propose a novel unified framework called \textit{ECSeq}, to introduce graph compression techniques into relation modeling for user sequence representation learning. The key module of \textit{ECSeq} is \textit{sequence relation modeling}, which explores relationships among sequences to enhance sequence representation learning, and employs graph compression algorithms to achieve high efficiency and scalability. \textit{ECSeq} also exhibits \textit{plug-and-play} characteristics, seamlessly augmenting pre-trained sequence representation models without modifications. Empirical experiments on both sequence classification and regression tasks demonstrate the effectiveness of \textit{ECSeq}. Specifically, with an additional training time of tens of seconds in total on 100,000+ sequences and inference time preserved within $10^{-4}$ seconds/sample, \textit{ECSeq} improves the prediction R@P$_{0.9}$ of the widely used LSTM by $\sim 5\%$.

\end{abstract}

\begin{IEEEkeywords}
Sequence Representation Learning, Graph Neural Network, Graph Compression, Online Inference
\end{IEEEkeywords}

\section{Introduction} 
\label{introduction}
\begin{figure}[tbp]
	\centering
	\includegraphics[width=0.9\linewidth]{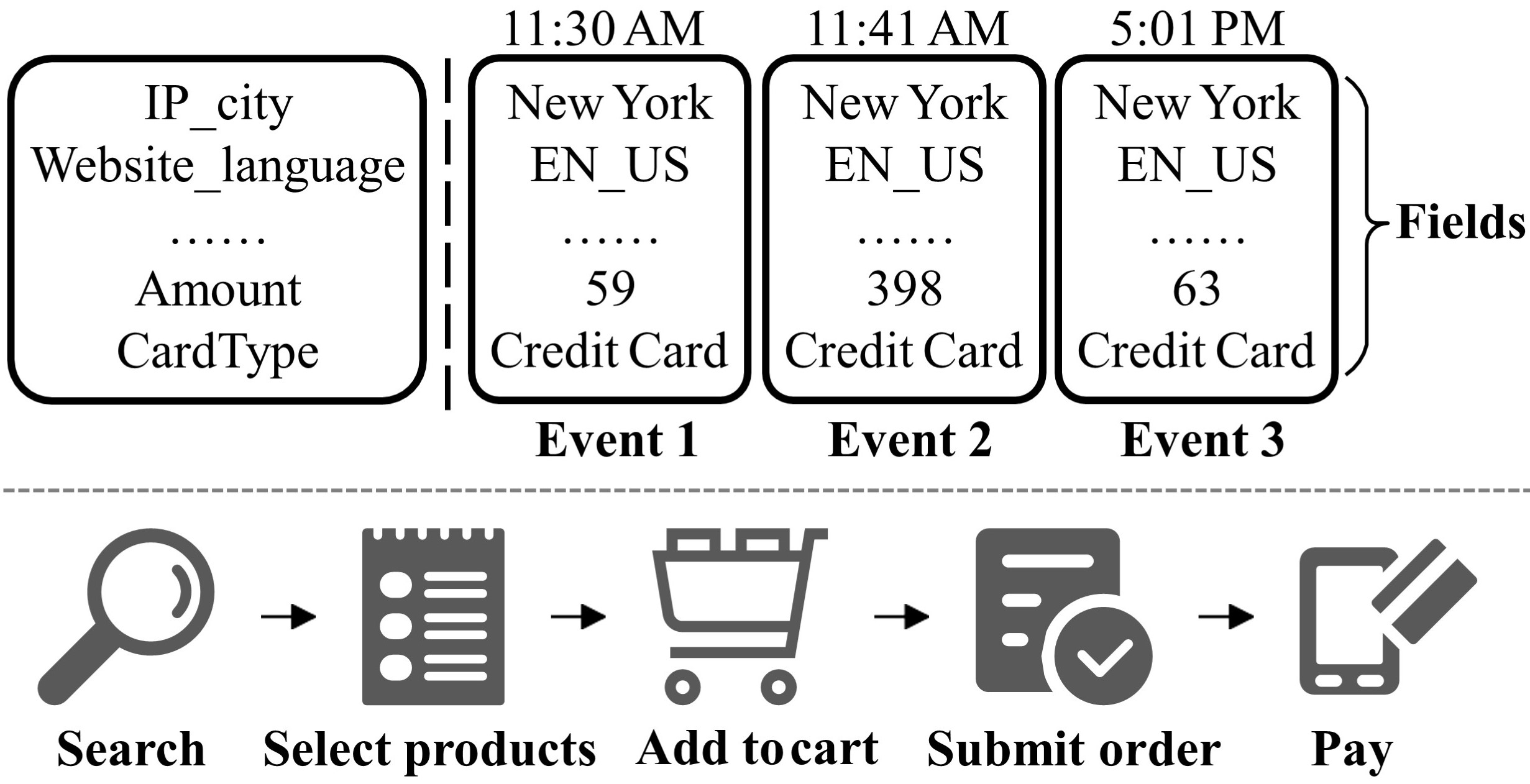}
	\caption{\textbf{Up:} A user behavior sequence example consists of three events and each event has several fields. \textbf{Down}: An example of user behavior sequence for online shopping.}
	\label{fig:seqdata}
 \vspace{-1em}
\end{figure}

User sequences data record user online activities over time. 
For example, online shopping users may exhibit sequences of behaviors such as search, add to cart, and payment, as depicted in Figure \ref{fig:seqdata}.
Learning effective user representations from sequence data is crucial for tasks like recommendation \cite{Hidasi_session_based_2015} and fraud detection \cite{wang_2017_pkdd,jurgovsky_sequence_2018}. Deep learning models like recurrent neural networks (RNNs)~\cite{jurgovsky_sequence_2018,wang_2017_pkdd}, convolutional neural networks (CNNs)~\cite{fu_credit_2016}, and self-attentive models~\cite{kang2018self,sun2019bert4rec} have shown promise for learning user sequence representations. 
These methods first map user sequences into embedding vectors, which subsequent layers leverage to conduct predictions.
In practice, relying solely on an individual user's historical behavior sequences is often insufficient for predicting future behavior, 
as it potentially overlooks information inherent in other pertinent behavioral sequences.
To address this, one feasible approach is to leverage similar sequences from other users and model their correlations~\cite{Zhang_Gao_Ma_Wang_Wang_Tang_2021,zhang2022m3care}.
Following this idea, many graph neural network (GNN) methods have been proposed to capture correlations among similar sequences for online user behavior modeling~\cite{liuIntention}.

However, most GNN-based sequence representation methods still face challenges with efficiency and scalability~\cite{huang2021scaling} for online services.
The number of user sequences produced by online applications is often immense, potentially escalating to the magnitude of millions~\cite{zhu_HEN}, which results in relation graphs with millions of nodes and edges. 
For GNN training, such large-scale graphs incur substantial computational and memory burdens, making model training difficult.
Furthermore, online services typically require rapid response, which puts stringent demands on the algorithms' inference efficiency. Consider the scenario of online shopping: the fraud detection mechanism must function in real time, completing just before the final payment, which ensures the timely interruption of potentially fraudulent transactions.
When deployed to online applications, existing methods (\textit{e.g.}, IHGAT~\cite{liuIntention}) need to compute similarities between each new sequence and all existing sequences, hardly meeting the requirement of low latency for inference.


To improve efficiency and scalability in both training and inference phases, we \textit{question the necessity of modeling all user sequences as nodes in GNNs}.
More specifically, although real-world applications involve vast sequence data, not every sequence may need representation in the graph.
For example, in fraud detection, many users exhibit similar purchasing sequences as in Figure~\ref{fig:seqdata} (search $\rightarrow$ select products $\rightarrow$ add to cart $\rightarrow$ submit order $\rightarrow$ pay). Thus, selecting a representative subset as nodes may suffice without compromising effectiveness.
That is, \textit{we can compress the graph by reducing nodes and edges prior to GNN model training, benefiting computational efficiency}.
Most existing GNNs like GCN~\cite{kipf2016semi} and GraphSAGE~\cite{HamiltonYL17} have time complexity quadratic in nodes or linear in edges~\cite{wu2020comprehensive}.
Hence, shrinking the graph size can directly accelerate computation. Additionally, graph compression can impart other desirable properties for user behavior sequence learning:

\begin{itemize} 
    \item \textit{Case-based Reasoning}~\cite{NIPS2014_390e9825}.
    Graph compression provides representative sequence prototypes~\cite{li2018deep,zhang2023learning}, offering interpretable cases of model outputs.

     \item \textit{Sample Balancing}.  
    In biased distributions (\textit{e.g.}, fraud detection) \cite{johnson2019survey,zhong2020financial}, compression can balance categories by setting similar compressed node counts.
\end{itemize}

Various \textit{graph compression} techniques exist in literature, including condensation~\cite{jin2022condensing,jin2022graph}, coarsening~\cite{huang2021scaling,cai2021graph}, and sparsification~\cite{peleg1989graph,spielman2011spectral}. However, which techniques suit user behavior sequence learning remains an open question.
Furthermore, as diverse GNN relational models like GCN~\cite{kipf2016semi}, GraphSAGE~\cite{HamiltonYL17}, and GAT~\cite{vaswani2017attention} may excel in different tasks, a unified representation learning framework that incorporates these varied techniques is desirable.

To this end, we design a unified framework, \textit{ECSeq}, to learn user behavior \textbf{Seq}uence representations \textbf{E}fficiently using \textbf{C}ompressed GNNs. 
Specifically, \textit{ECSeq} first extracts sequence information as nodes to construct a relation graph, then compresses it into a small synthetic graph. This compressed graph trains a GNN to generate sequence representations for downstream tasks, and the nodes in the compressed graph are perceived as representative sequences. 
During online inference, we incorporate new sequences as nodes into the compressed graph, then establish connections with representative sequences to enhance sequence representations through the message-passing process of GNN.

Our main contributions are summarized as follows:
\begin{itemize}
    \item We propose \textit{ECSeq}, a unified user sequence learning framework for online services, to incorporate the relations between target and similar sequences. \textit{ECSeq} enhances efficiency and scalability via graph compression, thus resolving the dilemma of relation modeling on large-scale sequence data and online inference with low latency.
    \item We compare and adapt suitable graph compression techniques for \textit{ECSeq}, meeting efficiency, interpretability, and sample balancing demands simultaneously. Besides, \textit{ECSeq} is designed to hold \textit{plug-and-play} characteristics, seamlessly augmenting pre-trained sequence representation models in existing systems without the need to modify these models.
    \item Experiments on transaction fraud detection (classification tasks) and user mobility (regression tasks) datasets demonstrate \textit{ECSeq}'s effectiveness, efficiency, and flexibility with different sequence learning, compression, and GNN options. 
    Specifically, with an extra training time of tens of seconds in total on 100,000+ sequences and inference time maintained within $10^{-4}$ seconds/sample, \textit{ECSeq} enhances the prediction R@P$_{0.9}$ of the widely used LSTM by $\sim 5\%$.
    Case studies also showcase \textit{ECSeq}'s interpretability.
    Codes and data are available at \href{https://github.com/wuyucheng2002/ECSeq}{https://github.com/wuyucheng2002/ECSeq}.
\end{itemize}

\section{Problem Formulation}

This section introduces key concepts and defines the problem.

\textbf{Field, Event, and Sequence.} 
User sequences comprise multiple events in chronological order. Events encapsulate user behaviors at a given moment or phase. Fields represent recorded feature values when an event occurs. For example, in online shopping scenarios (Figure \ref{fig:seqdata}), a user from New York paid \$59, \$398, \$63 for three orders at 11:30 AM, 11:41 AM, and 5:01 PM, respectively. The user paid by credit card with English language settings. Here, the three payment transactions are events, where the last transaction is the target event, while `IP\_city', `Website\_language', `Amount', and `CardType' are fields.

We assume the existence of $N$ sequences, represented by the set $E=\{E_1,E_2,...,E_N\}$, where $E_i$ is the $i$-th sequence and $1\leq i \leq N$. Each sequence $E_i = [e_{i}^{1}, e_{i}^{2}, ..., e_{i}^{T}]$, where $T$ is the number of events in each sequence and each event has $F_f$ fields. The $t$-th event in sequence $E_i$ is $e_{i}^{t}\in \mathbb R^{F_f}$, where $1\leq t \leq T$. $e_{i}^{T}$ is the target event, and its label is considered as the sequence label, which we need to predict. For categorical labels, $Y\in\{0,1\}^{N\times c}$ denotes the corresponding one-hot label of the sequence, where $c$ is the class number. For numerical labels, $Y\in\mathbb{R}^{N\times 1}$ represents the ground-truth value.

\textbf{Relation Graph.} 
The relation graph, denoted as $\mathcal{G} = (\mathcal{A}, \mathcal{X})$, is undirected and consists of an adjacency matrix $\mathcal{A} = \{a_{ij}\} \in \{0,1\}^{N \times N}$ and a node feature matrix $\mathcal{X} = \{x_i\} \in \mathbb{R}^{N \times D}$. Here, $N$ represents the number of nodes\footnote{We consider sequences as nodes when building the graph, so $N$ also denotes the number of sequences without ambiguity.}, $M$ denotes the number of edges, and $D$ indicates the dimension of node features. For classification and regression tasks, the node label matrix is given by $\mathcal{Y} \in \{0,1\}^{N \times c}$ and $\mathbb{R}^{N \times 1}$, respectively.

The \textit{compressed} relation graph is represented as $\tilde{\mathcal{G}} = (\tilde{\mathcal{A}}, \tilde{\mathcal{X}})$, where $\tilde{\mathcal{A}} \in \{0,1\}^{K \times K}$ and $\tilde{\mathcal{X}} \in \mathbb{R}^{K \times D}$. Here, $K$ is the number of nodes in the compressed graph, which is less than $N$. The corresponding node label matrix is denoted by $\tilde{\mathcal{Y}} \in \{0,1\}^{K \times c}$ or $\mathbb{R}^{K \times 1}$.

\textbf{User Sequential Learning Problem.} 
Given a sequence of user behavior $E_i = [e_i^1, e_i^2, ..., e_i^T]$, our task is to learn beneficial representation vectors, which are subsequently used to predict the target label (classification) or value (regression) of the sequence.\footnote{Note that the \textit{sample balancing} characteristic discussed in Section~\ref{introduction} is only required for classification tasks.
}

\section{The Proposed Framework}

\begin{figure*}[htbp]
	\centering
	\includegraphics[width=\textwidth]{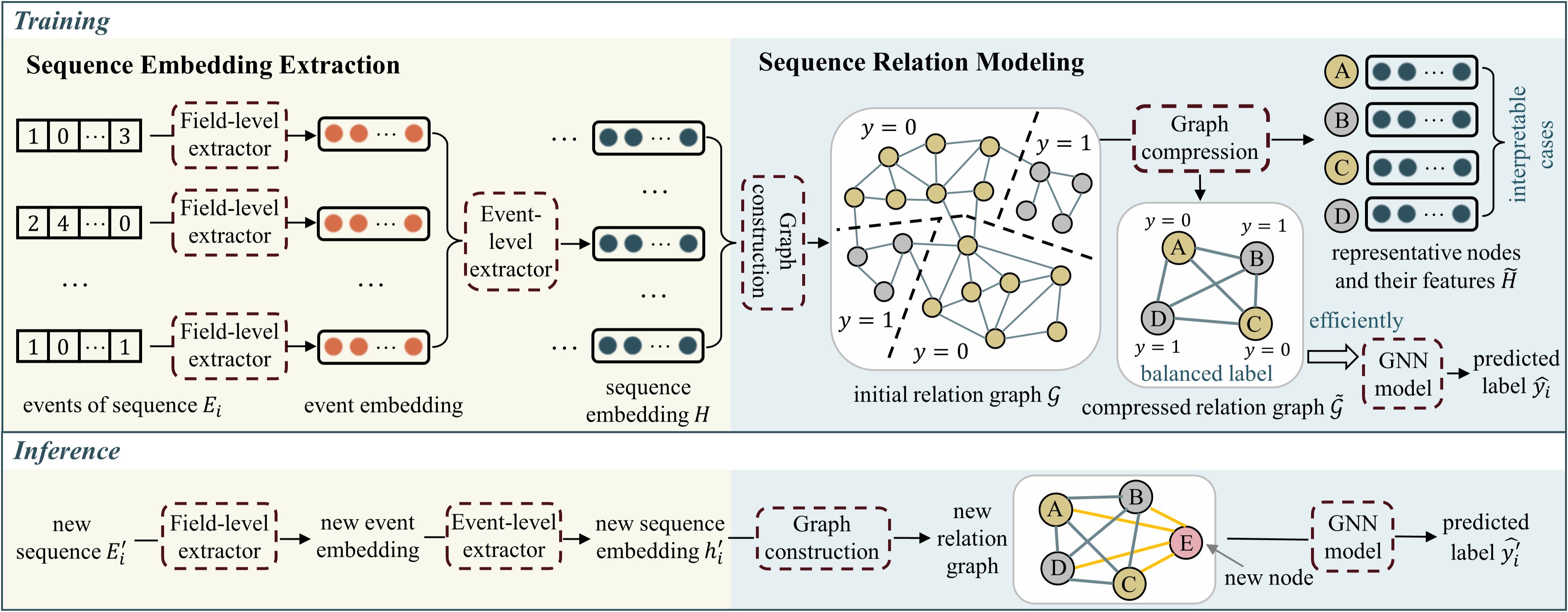}
	\caption{Overview of \textit{ECSeq}. Firstly, the \textit{sequence embedding extraction} module meticulously transforms sequence information into a one-dimensional feature vector. Then, the \textit{sequence relation modeling} module explores and leverages relationships among sequences to enhance the sequence representation, employing an appropriate graph compression technique to mitigate computational overhead and improve inference efficiency.}
	\label{fig: framework}
\end{figure*}

Figure~\ref{fig: framework} overviews \textit{ECSeq}.
The \textit{ECSeq} framework is comprised of two modules: \textit{sequence embedding extraction} and \textit{sequence relation modeling}. The former embeds sequence information into a one-dimensional feature vector, while the latter explores and leverages relationships among sequences to enhance the sequence representation. The efficiency of the relational model is amplified through the selection of appropriate graph compression techniques. The workflow of \textit{ECSeq} during the \textit{training} and \textit{inference} phases is also delineated in Figure~\ref{fig: framework}.

\subsection{Sequence Embedding Extraction}

The goal of sequence embedding extraction is to capture a user's sequence representation by taking into account both field-level characteristics and event-level sequential patterns over time.

The numerical data is normalized through max-min normalization, whereas categorical data is subjected to one-hot encoding in the field-level feature extractor. The event embedding $g_i^t$ is exported by the field-level extractor $\mathcal{M}_f$ as:
\begin{equation}
g_i^{t} = \mathcal M_f(e_i^{t})\in \mathbb{R}^{F_e}
\end{equation}

We utilize an event-level extractor, denoted as $ \mathcal M_e $, to extract temporal patterns from users' sequence data. The event-level extraction methods commonly use backbone networks that aggregate past and current messages such as CNN~\cite{tang2018personalized}, RNN~\cite{wang_2017_pkdd}, and attention-based models \cite{zhu_HEN}. The process of obtaining the event embedding $ h_i $ for event $ e_i^ { T } $ while considering historical sequences can be represented as:
\begin{equation}
h_i = \mathcal M_e(g_i^{1},g_i^{2},...,g_i^{T})\in \mathbb{R}^{F_s}
\end{equation}
$h_i$ can also represent the sequence embedding for $E_i$. 
The matrix $H \in \mathbb{R}^{N \times F_s}$ denotes all the sequence embeddings, where each row of $H$ corresponds to $h_i$.

Subsequent to $ \mathcal M_f $ and $ \mathcal M_e $, there comes a multi-layer perceptron (MLP) prediction head, denoted as $ \mathcal F_{seq} $, which takes the sequence embedding $ E $ as input and produces $ \hat Y $ as output. The output $ \hat Y $ is a soft label matrix of probability distribution vectors for categorical labels, represented as $ \hat Y \in \mathbb R^{N \times c}$, or a forecast value matrix for numerical labels, represented as $ \hat Y \in \mathbb R^{N \times 1}$.
\begin{equation}
\hat Y_i=\mathcal F_{seq}(h_i)
\end{equation}

We use the cross-entropy (CE) loss function for classification tasks and the mean squared error (MSE) loss function for regression tasks to effectively supervise the training process.
\begin{equation}
\begin{aligned}
\label{eq:loss_seq}
\textit{Classification:} \ &\mathcal L_{seq}=\textit{CE}\,(\hat Y,Y)=-\frac{1}{N} \sum_{i=1}^{N}\sum_{j=1}^{c}Y_{i,j} \log (\hat Y_{i,j})  \\
\textit{Regression:} \ &\mathcal L_{seq}=\textit{MSE}\,(\hat Y,Y)=-\frac{1}{N}  \sum_{i=1}^{N} (\hat Y_i-Y_i)^2
\end{aligned}
\end{equation}

Note that the design of field-level and event-level extractors is not the focus of this paper. Any new techniques proposed in future studies can be easily integrated into \textit{ECSeq}. In our experiments, we also validate that \textit{ECSeq} can incorporate different sequential feature extractors such as LSTM and transformers.

\subsection{Sequence Relation Modeling}
\label{sub:seq_relation_model}

\begin{figure*}[htbp]
\begin{minipage}{0.70\textwidth}
    \captionof{table}{Summary of typical graph compression methods. $N$: number of nodes, $M$: number of edges, $D$: dimension of node features, $K$: number of clusters/compressed nodes, $c$: some absolute constant. 
    \textit{Traceable}: whether the source of the compressed nodes is known; \textit{Configurable}: whether the compression method can assign separate compressed node quantities for each category.}
    \label{tab:graph_compression} 
	\centering
	\resizebox{\linewidth}{!}{
		\begin{tabular}{llcccccc}
            \toprule
            \multirow{2}{*}{\textbf{Category}} &\multirow{2}{*}{\textbf{Methods}} &\multirow{2}{*}{\textbf{Input}} &\multicolumn{2}{c}{\textit{Efficiency}} 
            &\textit{Interpretability} &\textit{Balancing} \\
            \cmidrule(lr){4-5} 
            \cmidrule(lr){6-6} \cmidrule(lr){7-7}
            & &  &\bf \#Nodes$\downarrow$  &\bf \#Edges$\downarrow$ 
            &\bf Traceable &\bf Configurable \\
            \midrule
            \multirow{4}{*}{Coreset Selection}  &$k$-means~\cite{jain2010data}  &$\mathcal X$  &\ding{51}  &\ding{51}  
            &\ding{51} &\ding{51} \\
            &AGC~\cite{zhang2019attributed}  &$\mathcal A, \mathcal X$  &\ding{51}  &\ding{51} 
            &\ding{51} &\ding{55} \\
            &Grain~\cite{zhang2021grain} &$\mathcal A, \mathcal X$   &\ding{51}  &\ding{51} 
            &\ding{51} &\ding{51}\\
            &VNG~\cite{si2023serving}  &$\mathcal A, \mathcal X$ &\ding{51}  &\ding{51} 
            &\ding{51}  &\ding{55} \\
            \midrule
            \multirow{3}{*}{Graph Coarsening}  
            &RSA~\cite{loukas2019graph}  &$\mathcal A$  &\ding{51} &\ding{51} 
            &\ding{51}  &\ding{55} \\
            &REC~\cite{loukas2018spectrally}  &$\mathcal A$  &\ding{51}  &\ding{51}  
            &\ding{51} &\ding{55} \\
            &GOREN~\cite{cai2021graph} &$\mathcal A$   &\ding{51}  &\ding{51} 
            &\ding{51}  &\ding{55} \\
            \midrule
            Graph Sparsification  &ApproxCut~\cite{spielman2011spectral}  &$\mathcal{A}$    &\ding{55} &\ding{51} 
            &\ding{51}  &\ding{55} \\
			\bottomrule
		\end{tabular}
  }
\end{minipage}
\quad
\begin{minipage}{0.27\textwidth}
	\centering
	\includegraphics[width=\linewidth]{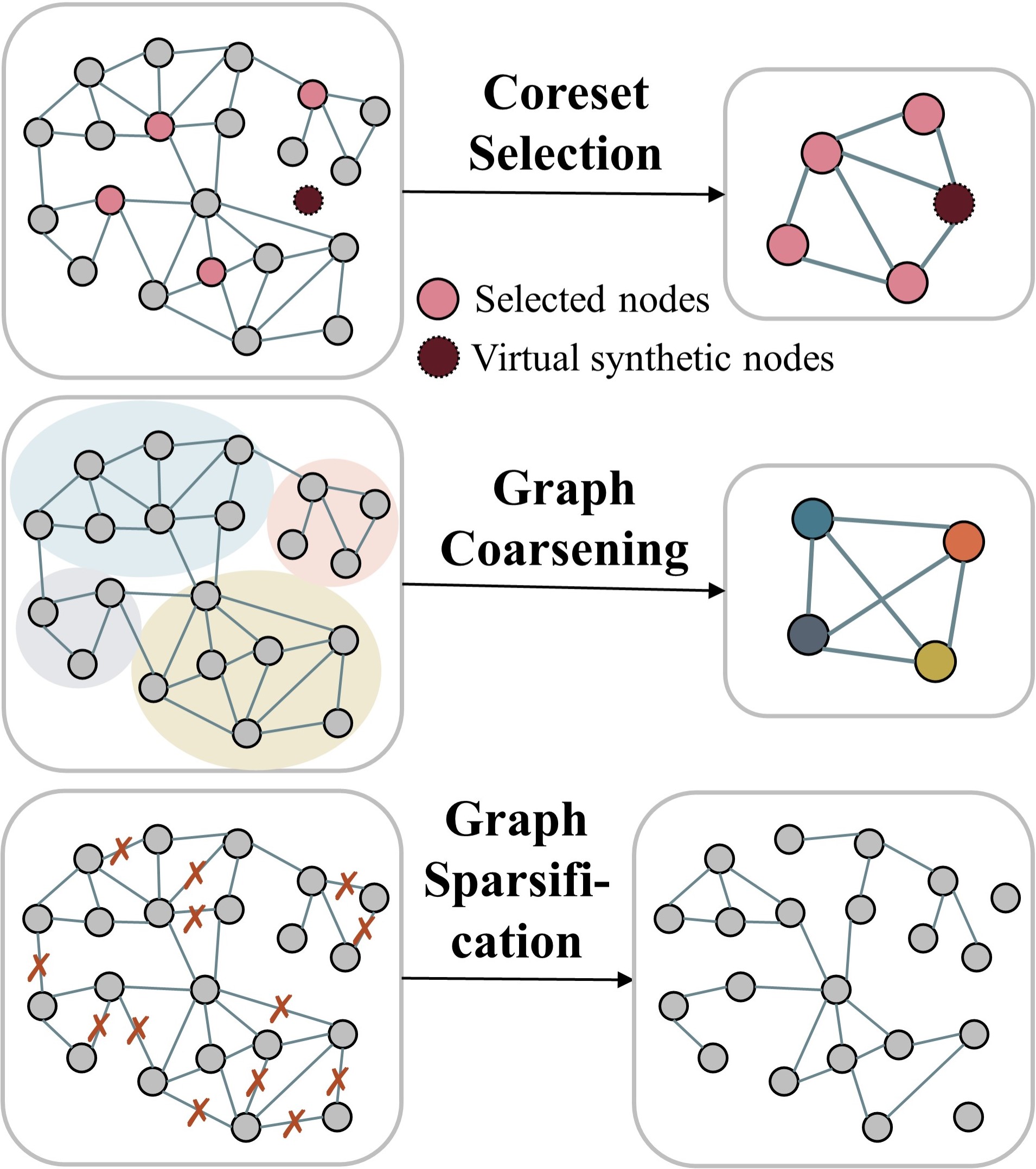}
	\caption{Illustration of diverse graph compression methods.}
	\label{fig:graph_compression}
\end{minipage}
\end{figure*}

One approach to leveraging the information of relevant users' sequences is to construct a relation graph based on the generated sequence embedding. However, directly acquiring node representation of the relation graph may result in inefficiency. Therefore, it is crucial to reduce the size of the graph using suitable graph compression techniques. Afterward, the GNN-based graph representation model can be trained on the small compressed graph to obtain node embeddings and predict node labels.

\subsubsection{\textbf{Graph Construction}}
\label{sec:graph_construction}

To begin, we regard users' sequences as nodes to construct a relationship graph, denoted as $ \mathcal G $. We express the feature matrix $ \mathcal X \in \mathbb R^ { N \times D } $ as the sequence embedding matrix $ H \in \mathbb R^ { N \times F_s } $, where $ \mathcal X = H $, $ D = F_s $, and $ \mathcal Y = Y $. However, the edges of $ \mathcal G $ are not known, and we can deduce them based on the available node attributes.

The homophily theory~\cite{mcpherson2001birds} suggests that connected nodes tend to share similar attributes. This allows us to infer node connectivity based on attribute similarity. There are various methods for constructing graphs, of which the two most prevalent are $k$-Nearest Neighbors ($k$-NN graphs)~\cite{preparata2012computational} and $\epsilon$-proximity thresholding ($\epsilon$-graphs)~\cite{Bentley1977TheCO}. Both methods calculate pairwise similarity scores of node features using similarity metrics. In $k$-NN graphs, two nodes $(v_i, v_j)$ are connected if $v_i$ has one of the $k$ highest similarity scores of $v_j$. In $\epsilon$-graphs, two nodes are linked if their similarity score is larger than a fixed threshold $\epsilon$. For example, using the similarity metric $sim$ (\textit{e.g.}, cosine similarity) in $\epsilon$-graphs, edges can be created as follows:

\begin{equation}
\mathcal A_{ij}=\left\{
    \begin{aligned}
    &1,\, sim(x_i,x_j)>\epsilon \\
    &0,\, sim(x_i,x_j)\leq \epsilon
    \end{aligned}
\right.
,\; 1\leq i,j\leq N
\end{equation}

Besides, we can enhance model performance by incorporating domain-specific knowledge for particular tasks into the construction of graphs, such as sequence feature co-occurrence~\cite{liang_sigir_2019}, user's social relationships~\cite{jiangAnomaly2019}, location functionality similarity~\cite{geng2019spatiotemporal}, and geographic proximity~\cite{wang2021exploring}. 
In particular, to consider social relationships, we can establish connections among behavior sequences whose users have established friendships.
When encompassing geographic proximity knowledge, we can calculate the geographical distance using the nodes' latitudinal and longitudinal data, linking the close instances. 
As for location functionality similarity, we deduce the nodes' functionality from their historical records, and nodes exhibiting similar historical behavior patterns will establish connections.

\subsubsection{\textbf{Graph Compression}} 
Our approach to addressing the efficiency issue involves adopting graph compression techniques that also achieve interpretability and sample balancing objectives. To ensure that our requirements are met, we conduct a thorough comparison and analysis of various graph compression algorithms before selecting the most suitable ones.

In GNN, graph compression aims to reduce the number of nodes or edges while maintaining model performance. Representative approaches can be classified into the following categories:\footnote{\textit{Graph condensation}~\cite{jin2022graph,jin2022condensing} and \textit{pooling}~\cite{zhang2018end,ying2018hierarchical} also generate synthetic graphs with reduced size. However, the nodes in the output graph lack practical meaning as node features become unexplainable latent vectors. This makes it difficult to define connections between new nodes and synthetic nodes during the inference phase. Hence, these techniques may not be suitable for inductive inference on new user sequences.}

\begin{itemize} 

    \item \textit{Coreset Selection} expedites training by selecting or synthesizing a subset of representative nodes. Grain~\cite{zhang2021grain} connects data selection in GNN with social influence maximization. VNG~\cite{si2023serving} constructs a small set of virtual nodes based on the GNN forward pass reconstruct error. Node clustering methods, such as $k$-means~\cite{jain2010data}, can also be considered as coreset selection, as clustering centroids can be selected as representative nodes. Some clustering methods are specifically designed for GNN, such as AGC~\cite{zhang2019attributed}, which performs spectral clustering based on nodes' hidden features in GNN.

    \item \textit{Graph Coarsening} 
    combines original nodes into super-nodes and establishes their connections~\cite{loukas2018spectrally,loukas2019graph,deng2019graphzoom,huang2021scaling}. Different from coreset selection, it usually only considers edges (\textit{i.e.}, structural information). Loukas proposes a graph coarsening method based on restricted spectral approximation (RSA)~\cite{loukas2019graph}, while GOREN~\cite{cai2021graph} is an unsupervised GNN-based framework to learn edge weights.    

    \item \textit{Graph Sparsification} (a.k.a. pruning) reduces the number of edges in a graph by approximating its structural properties, such as distances~\cite{peleg1989graph}, cuts~\cite{karger1994random} or eigenvalues~\cite{spielman2011spectral}. Note that all the original nodes are kept in the sparsified graph.
    
\end{itemize}

Table~\ref{tab:graph_compression} summarizes common graph compression methods that can significantly reduce the number of nodes ($N$) and/or edges ($M$), and Figure~\ref{fig:graph_compression} illustrates the principles of various categories of these methods. These methods can speed up the training and inference process, which is especially useful since most GNN models have a complexity of $\mathcal{O}(N^2)$ or $\mathcal{O}(M)$ \cite{wu2020comprehensive}. Although many compression methods listed in Table \ref{tab:graph_compression} create synthetic nodes, it is still possible to identify the original nodes used to generate these synthetic nodes. This provides the opportunity for case-based reasoning interpretability by tracing back to the original nodes.

\textbf{Sample Balancing}. 
To address the problem of sample imbalance, we can modify current techniques to ensure that the compressed graph contains a similar number of nodes with different labels.
\begin{itemize} 
    \item For traditional clustering methods like $k$-means~\cite{jain2010data}, we can cluster nodes in different categories separately and set a similar clustering number for each category. With Grain~\cite{zhang2021grain}, which employs a greedy node selection algorithm, we can terminate the search loop when the number of selected nodes in each category reaches a certain value. 

    \item However, for other techniques in Table~\ref{tab:graph_compression} relying on global graph properties (\textit{e.g.} eigenvalues~\cite{spielman2011spectral,loukas2018spectrally}, Laplace operator~\cite{cai2021graph}), the number of compressed nodes in each category is hard to set freely.  This makes it challenging to balance compressed nodes with different labels.
\end{itemize}

In summary, node clustering techniques, \textit{e.g.}, $k$-means (coreset selection) can achieve efficiency, interpretability, and sample balancing simultaneously, as shown in Table~\ref{tab:graph_compression}. 
Meanwhile, if certain properties are unnecessary (\textit{e.g.}, regression tasks do not need sample balancing), other methods like AGC/VNG (coreset selection) and RSA/REC/GOREN (coarsening) can also be considered.

With graph compression techniques, we can derive a new graph denoted as $ \tilde { \mathcal G } $. To represent the node mapping function between $ \mathcal G $ and $ \tilde { \mathcal G } $, we define an assignment matrix $ \Phi \in \mathbb R^ { N \times K } $. Specifically, $ \Phi_{ij} $ indicates the amount of information from the $i$-th node in $ \mathcal G $ that contributes to the $ j $-th node in $ \tilde { \mathcal G } $. For instance, in the clustering method, the $ j $-th node in $ \tilde { \mathcal G } $ represents the centroid of the $ j $-th cluster $ \mathcal C_j $, where $ \mathcal C_j $ is a node set of $ \mathcal G $ and $ | \mathcal C_j| $ denotes its count. If the centroid's feature vector is set by calculating the average of all the feature vectors of nodes in $ \mathcal C_j $, then $\Phi_{ij}$ can be defined as:
\begin{equation}
    \Phi_{ij}=\left\{
    \begin{aligned}
    &\frac{1}{|\mathcal C_j|},\, v_i\in \mathcal C_j \\
    &0,\, v_i\not\in\mathcal C_j
    \end{aligned}
\right.
,\; 1\leq i\leq N,\; 1\leq j\leq K
\end{equation}
When using compression methods that select real nodes (\textit{e.g.}, Grain \cite{zhang2021grain}), if the $j$-th node in $\tilde {\mathcal G}$ corresponds to the $i$-th node in $\mathcal G$, then $\Phi_{ij}=1$, otherwise $\Phi_{ij}=0$. 
$\Phi$ thus acts as a connector between the sequence embedding extractor and the GNN model. We can obtain node features $\tilde {\mathcal X}$ and node labels $\tilde {\mathcal Y}$ by:
\begin{equation}
\label{eq:assignment}
    \tilde {\mathcal X} = \Phi^{\mathrm{T}}\mathcal X,\; \tilde {\mathcal Y} = \Phi^{\mathrm{T}}\mathcal Y
\end{equation}
The compressed nodes are considered as representative sequences and their embedding is denoted by $\tilde{H} = \tilde{\mathcal{X}} \in \mathbb{R}^{K \times D}$, where $K$ is the number of compressed nodes (\textit{i.e.}, representative sequences).
$\Phi$ also records the source of representative sequences, serving as a reference for subsequent prediction, which would be useful for case-based reasoning.

\subsubsection{\textbf{Graph Prediction}}

To leverage the sequence relationship, we first extract node representations of the compressed relation graph $\tilde{\mathcal{G}}$ using state-of-the-art GNN models like GCN~\cite{kipf2016semi}, GraphSAGE~\cite{HamiltonYL17}, and GAT~\cite{vaswani2017attention}. The node representation matrix $\tilde W$ is generated as follows:
\begin{equation}
\tilde W = \mathcal M_g(\tilde{\mathcal{A}}, \tilde{\mathcal{X}}) \in\mathbb{R}^{K\times F_w}
\end{equation}
where $\mathcal M_g$ denotes the GNN model, and $F_w$ is the dimension of the node representation after graph neural layers. 
The final prediction is given through an MLP prediction head $\mathcal F_{gnn}$:
\begin{equation}
\hat{\mathcal Y} =\mathcal F_{gnn}(\tilde W)
\end{equation}
For classification, $\hat{\mathcal Y}\in\mathbb R^{K\times c}$ is the predicted soft label matrix, while for regression $\hat{\mathcal Y}\in\mathbb R^{K\times 1}$ contains predicted values. $\tilde{\mathcal Y}$ represents ground-truth labels. Note for classification, $\tilde{\mathcal Y}$ may not be one-hot after Eq.~\ref{eq:assignment} compression (\textit{e.g.}  samples with different labels exist in one cluster). If $\tilde{\mathcal Y}$ remains one-hot, we use cross-entropy loss; otherwise, mean squared error (MSE) is used. Regression always uses MSE loss.
\begin{equation}
\begin{aligned}
\label{eq:loss_com}
\textit{Classification:}\ 
&\mathcal L_{com}=\left\{
    \begin{aligned} 
    &\textit{CE}\,(\hat{\mathcal Y},\tilde{\mathcal Y}),
    \,  \tilde{\mathcal Y}\in\{0,1\}^{K\times c} \\
    &\textit{MSE}\,(\hat{\mathcal Y},\tilde{\mathcal Y}),
    \,  \tilde{\mathcal Y}\not\in\{0,1\}^{K\times c}\\
    \end{aligned}
\right.\\
\textit{Regression:} \ &\mathcal L_{com}=\textit{MSE}\,(\hat{\mathcal Y},\tilde{\mathcal Y})
\end{aligned}
\end{equation}

Training only on the compressed graph may be insufficient, as the model may not seamlessly generalize to new real sequences connected to existing compressed nodes.
To address this, we also train on correlations between original real nodes and compressed nodes, bridging the gap.
Specifically, we connect original nodes $\mathcal X$ (divided into batches if $N$ is large) to compressed nodes $\tilde{\mathcal X}$ (with methods in Section~\ref{sec:graph_construction}), forming a new adjacency matrix $\mathcal A'$.
We then aggregate information passing from compressed to real nodes via $\mathcal M_g$, making predictions with $\mathcal F_{gnn}$:
\begin{equation}
    W'=\mathcal M_g(\mathcal A', \mathcal X\cup\tilde{\mathcal X}),\;
    \hat Y'=\mathcal F_{gnn}(W')
\end{equation}

We optimize the model with the loss function similar to Eq.~\ref{eq:loss_seq}:
\begin{equation}
\begin{aligned}
\label{eq:loss_cor}
\textit{Classification:} \ &\mathcal L_{cor}=\textit{CE}\,(\hat Y',Y)
\\
\textit{Regression:} \ &\mathcal L_{cor}=\textit{MSE}\,(\hat Y',Y)
\end{aligned}
\end{equation}

\subsection{The Overall Workflow}

\textit{ECSeq} conducts sequence classification or regression for online services in two phases: training (Algorithm~\ref{alg:ecseq_train}) and inference (Algorithm~\ref{alg:ecseq_infer}).


\subsubsection{\textbf{Training}}

Jointly optimizing the \textit{sequence embedding extractor} (field-level extractor $\mathcal M_f$, event-level extractor $\mathcal M_e$, and prediction head $\mathcal F_{seq}$) and \textit{relation model} (GNN model $\mathcal M_{g}$ and prediction head $\mathcal F_{gnn}$) may be unstable. Thus we train them independently as follows:
\begin{itemize} 
    \item \textit{Step 1}. We use $\mathcal L_{seq}$ in Eq.~\ref{eq:loss_seq} as the loss function to train the embedding extractor to obtain sequence embeddings.

    \item \textit{Step 2}. We compress the graph and derive the assignment matrix $\Phi$. 

    \item \textit{Step 3}. We use $\mathcal L_{com}$ in Eq.~\ref{eq:loss_com} as the loss function to train the relation model on the compressed graph.
\end{itemize}

Our \textit{step-wise} approach has advantages over end-to-end training: (1) Graph compression is only needed once, mitigating instability and inefficiency; (2) Both modules' training processes incorporate label information supervision, ensuring the models' stability and optimality; (3) Decoupled optimization enables flexibility in using diverse models without alignment concerns.

Importantly, \textit{ECSeq}'s step-wise training enables seamlessly enhancing any pre-trained sequence representation model by using it as the embedding extractor, without modification (\textit{Step 1} can then be skipped).
This provides a \textit{plug-and-play} capability to augment relational modeling in current sequential learning systems. Thus, \textit{ECSeq} can readily be deployed in practice to boost existing models.

Subsequently, we refine the relation model to improve its generalization to new real sequences (nodes). Specifically, we employ graph construction methods (Section~\ref{sec:graph_construction}) to build connections between real and compressed nodes. Afterward, we aggregate real nodes' information transmitted through the compressed nodes and make predictions. The relation model is further optimized on the objective $\mathcal L_{cor}$ in Eq.~\ref{eq:loss_cor}. 

\subsubsection{\textbf{Inference}}
During the process of inference, we incorporate information from small-scale representative sequences into new sequences. To start, we extract information from the new sequence and obtain its sequence embedding. We then establish connections between the new sequence and the compressed relation graph to gain insights from the representative nodes (sequences) with methods in Section~\ref{sec:graph_construction}. Finally, we generate the representation of the new sequence and derive the prediction result.

\begin{algorithm2e}[t]
	\caption{\textit{ECSeq} Training Procedure}
 \label{alg:ecseq_train}
	\LinesNumbered 
	\KwIn{Sequence set $E$ and its label matrix $Y$}
	\KwOut{Optimized sequence embedding extractor ($\mathcal M_f$, $\mathcal M_e$), optimized relation model ($\mathcal M_{g}$, $\mathcal F_{gnn}$), and the compressed graph $\tilde{\mathcal G}$.}
    Initialize parameters of sequence embedding extractor $\mathcal M_f$, $\mathcal M_e$, and $\mathcal F_{seq}$\;
	\While{stopping condition is not met}{
		$H=\mathcal M_e(\mathcal M_f(E))$, $\hat Y=\mathcal F_{seq}(H)$\;
        Compute the loss $\mathcal L_{seq}$ by Eq.~\ref{eq:loss_seq}\;
        Update the parameters of $\mathcal M_f$, $\mathcal M_e$, and $\mathcal F_{seq}$\;
	}

    Treat sequences as nodes with $\mathcal X=H$\;
    Construct node connections and get relation graph $\mathcal G=(\mathcal A,\mathcal X)$\;
    Compress $\mathcal G$ to get the compressed graph $\tilde{\mathcal G}=(\tilde{\mathcal A}, \tilde{\mathcal X})$ and label $\tilde{\mathcal Y}$\;
    Initialize parameters of relation model $\mathcal M_g$ and $\mathcal F_{gnn}$\;
	\While{stopping condition is not met}{
		$\hat{\mathcal{Y}}=\mathcal F_{gnn}(\mathcal M_g(\tilde{\mathcal A}, \tilde{\mathcal X}))$\;
        Compute the loss $\mathcal L_{com}$ by Eq.~\ref{eq:loss_com}\;
        Update the parameters of $\mathcal M_g$ and $\mathcal F_{gnn}$\;
	}
 
    Establish connections between $\mathcal X$ and $\tilde{\mathcal X}$, denoted as $\mathcal A'$\;
    \While{stopping condition is not met}{
		$\hat Y'=\mathcal F_{gnn}(\mathcal M_g(\mathcal A', \mathcal X\cup\tilde{\mathcal X}))$\;
        Compute the loss $\mathcal L_{cor}$ by Eq.~\ref{eq:loss_cor}\;
        Update the parameters of $\mathcal M_g$ and $\mathcal F_{gnn}$\;
	}
\end{algorithm2e}

\begin{algorithm2e}[t]
	\caption{\textit{ECSeq} Inference Procedure}
 \label{alg:ecseq_infer}
	\LinesNumbered 
	\KwIn{Optimized sequence embedding extractor ($\mathcal M_f$, $\mathcal M_e$), optimized relation model ($\mathcal M_{g}$, $\mathcal F_{gnn}$), the compressed graph $\tilde{\mathcal G}$, and a set of new sequences $E''$}
	\KwOut{The predicted label $\hat Y''$ of the new sequences.}
    Get new sequence embedding $H''=\mathcal M_e(\mathcal M_f(E''))$\;
    Treat new sequences as nodes with $\mathcal X''=H''$\;
    Establish connections between $\mathcal X''$ and $\tilde{\mathcal X}$, denoted as $\mathcal A''$\;
    Derive inference results $\hat Y''=\mathcal F_{gnn}(\mathcal M_g(\mathcal A'', \mathcal X''\cup\tilde{\mathcal X}))$\;
\end{algorithm2e}

\section{Experiments}
\subsection{Experimental Settings}
\subsubsection{\textbf{Datasets}} \label{datasetdescription}
We use four datasets in two different scenarios. The details of the datasets are listed in Table~\ref{tab:datasetsStatistics}.

(1) \textit{Fraud detection} (binary classification task). 
The datasets, \textit{FD1} and \textit{FD2}, have been collected from a global e-commerce company. These datasets consist of real-world online card transaction sequences from two platforms, with all identifiable information removed. The objective is to detect fraudulent behavior sequences. The training, validation, and testing phases are conducted using the first four months, fifth month, and sixth month transaction data, respectively.

(2) \textit{User Mobility} (regression task). 
The \textit{Bike} dataset was collected from open data portals in New York City\footnote{\href{https://www.citibikenyc.com/system-data}{https://www.citibikenyc.com/system-data}}. Our study focuses on forecasting the number of bike-sharing demands at each station (\textit{i.e.}, the number of bike borrowers)~\cite{wang2021exploring}. The \textit{Speed} dataset (PEMS-BAY) contains traffic speed data from the Bay Area, with sensors that record highway vehicle speeds. Each sensor can be considered a node, and we predict the traffic speed of each sensor for the next time step. We model the temporal features of both datasets following~\cite{wang2021exploring}, and split each dataset into training/validation/testing sets in chronological order with a proportion of 8:1:1.

\begin{table}[t]
	\caption{Statistics of datasets.}\label{tab:datasetsStatistics}
	\begin{center}
    \begin{adjustbox}{max width=1\linewidth}
		\begin{tabular}{lcccc}
			\toprule
            \multicolumn{5}{c}{\textbf{\textit{Fraud Detection}}} \\
            Datasets &\#Fields &\#Events &\#Sequences &\#Positive Samples \\
            \midrule 
            \textbf{FD1} & 236 & 2,130,962 & 245,045 & 24,489 (9.99\%) \\
            \textbf{FD2} & 178 & 275,322 & 15,366  & 777 (5.06\%) \\
            \toprule 
            \multicolumn{5}{c}{\textbf{\textit{User Mobility}}} \\
            Datasets &\#Sensors  &\#Timesteps & Time Interval &Value Range \\
            \midrule 
            \textbf{Bike} &717 &1,488 &60 minutes &0.0 $\sim$ 108.0\\
            \textbf{Speed} &325 &1,488 &60 minutes  &3.1 $\sim$ 83.2\\
			\bottomrule
		\end{tabular}
    \end{adjustbox}
	\end{center}
    \vspace{-2em}
\end{table}

\subsubsection{\textbf{Metrics}}  

(1) For fraud detection, we adopt two widely-used metrics, AUPRC (Area Under the Precision-Recall Curve) and $\text{R@P}_{N}$ (Recall when Precision equals $N$).  
Considering the generally low rate of positive samples in fraud detection scenarios (less than 10\% in our datasets), AUPRC is suitable for evaluating highly imbalanced and skewed datasets~\cite{davis2006relationship,ding2023cross}.
Besides, we employ $\text{R@P}_N$ with $N=0.9$ following the previous work~\cite{liuIntention}. This metric increases more than nine-fold in our dataset (90\% vs 9.99\%/5.06\%), which underscores the capability to detect top-ranked positive samples without severely harming legitimate ones, thereby balancing the impact on real-world applications.
Higher AUPRC and $\text{R@P}_{0.9}$ indicate better performance. 

(2) For user mobility, we use RMSE (Root Mean Square Error)~\cite{wang2021exploring} and sMAPE (symmetric Mean Absolute Percentage Error)~\cite{makridakis1993accuracy}.
RMSE reflects the deviation between the predicted value and the true value, whereas sMAPE considers the ratio between the error and the true value. 
They are calculated as follows:
\begin{equation}
    \text{RMSE}=\sqrt{\frac{1}{N}\sum_{i=1}^{N}(Y_i-\hat Y_i)^2},\; 
    \text{sMAPE}=\frac{1}{N}\sum_{i=1}^{N}\frac{|Y_i-\hat Y_i|}{(Y_i+\hat Y_i)/2}
\end{equation}
where $Y_i$ means the true value of sample $i$, and $\hat Y_i$ indicates its predicted value.
A decline in RMSE and sMAPE signals enhanced performance.

\subsubsection{\textbf{Baselines}}
We employ five sequence representation learning methods in three categories as baselines.

\begin{itemize} 
    \item \textbf{Methods with only features of the target event}: \textit{Regression} (Logistic Regression for classification task and Ridge Regression for regression task) and \textit{GBDT} (Gradient Boosting Decision Tree)
    take features extracted by the field-level extractor of the target event as inputs to train a machine-learning classifier.
 
    \item \textbf{Methods with deep neural networks to extract historical information}: \textit{LSTM}~\cite{LSTMHochreiter} can capture long-term dependencies for sequential data, then we give the prediction by MLP layers, while \textit{R-Transformer}~\cite{wang2019r} combines RNNs and the multi-head attention mechanism.

    \item \textbf{Methods with GNN to capture sequence relationship}: \textit{GRASP}~\cite{Zhang_Gao_Ma_Wang_Wang_Tang_2021} enhances representation learning by leveraging knowledge extracted from similar users within the same batch, which is originally proposed for healthcare sequence classification problems. 
\end{itemize}

\subsubsection{\textbf{Implementations}}

In the implementation of \textit{ECSeq}, we adopt LSTM~\cite{LSTMHochreiter} to extract sequence embedding following existing work~\cite{branco2020interleaved,jurgovsky_sequence_2018}; we also test  \textit{ECSeq}'s flexibility of using other sequence embedding extractor, such as R-Transformer~\cite{wang2019r}.
We employ $\epsilon$-graphs to construct the relation graph, using sequence embedding (for fraud detection datasets) or location functionality~\cite{wang2021exploring} (for user mobility datasets) as the similarity feature.
We employ $k$-means~\cite{jain2010data} (default), AGC~\cite{zhang2019attributed}, Grain~\cite{zhang2021grain}, and RSA~\cite{loukas2019graph} to conduct graph compression. 
And we utilize GCN~\cite{kipf2016semi}, GraphSAGE~\cite{HamiltonYL17} (default), and GAT~\cite{velickovic2018graph} as GNN model with a 1-layer convolutional layer for graph representation learning.

\begin{table}[t]
	\caption{Parameter settings of \textit{ECSeq}. CN: compressed nodes; HL: hidden layers.}
 \label{tab:parameter_settings}
	\begin{center}
    \begin{adjustbox}{max width=1\linewidth}
		\begin{tabular}{lccccc}
			\toprule
            \multirow{2}[2]{*}{\textbf{Parameters}} &\multicolumn{2}{c}{\textit{\textbf{Fraud Detection}}} &\multicolumn{2}{c}{\textit{\textbf{User Mobility}}} \\
            \cmidrule(lr){2-3}	\cmidrule(lr){4-5}
            &\quad\textbf{FD1} &\textbf{FD2} &\quad\textbf{Bike} &\textbf{Speed} \\
            \midrule 
            Similarity feature &\multicolumn{2}{c}{Sequence embedding} &\multicolumn{2}{c}{Location functionality}  \\
            Similarity metric &\multicolumn{2}{c}{Cosine similarity} &\multicolumn{2}{c}{Pearson correlation}\\
            $\epsilon$ ($\epsilon$-graphs)  &\quad0.95  &0.99 &\quad0.5  &0.6 \\
            $k$ ($k$-means) / \#CN  &\quad500  &500  &\quad100  &100\\
            Positive CN ratio &\quad0.3  &0.2  &\quad--  &--\\
            \#HL of LSTM   &\quad256  &256 &\quad64  &64\\
            \#HL of GNN  &\quad32  &32  &\quad16 &16\\
		\bottomrule
		\end{tabular}
    \end{adjustbox}
	\end{center}
    \vspace{-2em}
\end{table}

We choose Adam optimizer with learning rate of 0.00001 for LSTM and with learning rate of 0.005 for GNN, and the batch size is 64.
The detailed parameter settings for each dataset are presented in Table~\ref{tab:parameter_settings}.
We use early-stopping to avoid overfitting. 
We repeat each experiment 5 times and report the mean and standard deviation.

\subsubsection{\textbf{Hardware}} 
Our experiment platform is a computation server with AMD Ryzer 9 3900X CPU (12 cores @ 3.80GHz), 64 GB RAM, and NVIDIA RTX 2080Ti GPU (11GB).

\subsection{Experimental Results}

\subsubsection{\textbf{Effectiveness}}

\begin{table*}[htbp]
	\caption{Experimental results on fraud detection and user mobility tasks. The best results are highlighted in bold. While \textit{R-Transformer}~\cite{wang2019r} and \textit{GRASP}~\cite{Zhang_Gao_Ma_Wang_Wang_Tang_2021} are primarily intended for classification tasks, they do not show comparable performance on user mobility tasks.} 
	\label{tab:results}
	\centering
    \begin{adjustbox}{max width=1\linewidth}
		\begin{tabular}{lccccccccccc}
		\toprule
        \multirow{2}[2]{*}{\textbf{Methods}} &\multicolumn{2}{c}{\textbf{FD1}} & \multicolumn{2}{c} {\textbf{FD2}} & \multicolumn{2}{c}{\textbf{Bike}} & \multicolumn{2}{c}{\textbf{Speed}}\\ 
        \cmidrule(lr){2-3} \cmidrule(lr){4-5} \cmidrule(lr){6-7} \cmidrule(lr){8-9}
        & AUPRC  ($\uparrow$) & $\text{R@P}_{0.9}$ ($\uparrow$) & AUPRC  ($\uparrow$) & $\text{R@P}_{0.9}$ ($\uparrow$) & RMSE ($\downarrow$) &sMAPE ($\downarrow$) &RMSE ($\downarrow$) &sMAPE ($\downarrow$)  \\
        \midrule
        \multicolumn{5}{l}{\textbf{Non-Graph Methods}}\\ 

        \textit{Regression} &0.7685±0.0000 &0.4890±0.0000 &0.5271±0.0000 &0.3052±0.0000 &2.8169±0.0000 &0.2148±0.0000 &6.3994±0.0000 &0.0672±0.0000 \\ 
        \textit{GBDT} &0.7742±0.0000 &0.5244±0.0000 &0.6147±0.0006 &0.3766±0.0000 &2.7596±0.0077 &0.2063±0.0008 &6.2964±0.0575 &0.0632±0.0005 \\
        \textit{LSTM} &0.8332±0.0047 &0.5840±0.0132 &0.7124±0.0076 &0.6987±0.0078 &1.5463±0.0504 &0.1782±0.0040 &4.8383±0.1600 &0.0561±0.0011\\
        \textit{R-Transformer} &0.8338±0.0040 &0.5847±0.0289 &0.7064±0.0149 &0.5403±0.1951  &-- &-- &-- &--\\
        \midrule
        \multicolumn{5}{l}{\textbf{Graph Methods}}\\
        \textit{GRASP} &0.8362±0.0037 &0.6049±0.0230 &0.7138±0.0353 &0.6776±0.0420 &-- &-- &-- &--\\
        \textit{ECSeq} 
        &\textbf{0.8383±0.0018}
        &\textbf{0.6153±0.0079}
        &\textbf{0.7249±0.0112} &\textbf{0.7039±0.0032}
        &\textbf{1.4832±0.0209} &\textbf{0.1766±0.0038} &\textbf{4.4362±0.1015} &\textbf{0.0542±0.0012} \\
        
        \bottomrule 
	\end{tabular}
 \end{adjustbox}
\end{table*}

Table~\ref{tab:results} represents the fraud detection and user mobility results of \textit{ECSeq} and baseline methods. Compared to \textit{Regression} and \textit{GBDT} that only use features of target event, \textit{LSTM} and \textit{R-Transformer} extract historical information of sequences, and have significant improvement in all datasets. \textit{ECSeq} further enhances \textit{LSTM} by exploring and utilizing the relationships among sequences, and \textit{ECSeq} gains the best performance in all of the four datasets. Besides, since \textit{GRASP} only constructs graphs within sequence batches, graphs may only capture relations among a small scale of samples. \textit{ECSeq} considers the sequence relation from a global perspective and is invulnerable to the randomness of batch partitioning, which reveals higher performance and stability.

\subsubsection{\textbf{Efficiency and Scalability}}

\begin{table*}[htbp]
\centering 
	\caption{Fraud detection performance and computation time on \textit{FD1}, whose training set contains 145,236 sequences, \textit{i.e.}, 145,236 original nodes when modeling the relation. We train the GNN for 50 epochs.
 } 
	\label{tab:scalability}
    \begin{adjustbox}{max width=\linewidth}
		\begin{tabular}{lccccccccccc}
		\toprule
        \textbf{Methods} &\#Nodes
        &\makecell{AUPRC \\($\uparrow$)} 
        &\makecell{$\text{R@P}_{0.9}$ \\($\uparrow$)} &\makecell{Compression \\Time (s) ($\downarrow$)} &\makecell{GNN Training \\Time (s) ($\downarrow$)} &\makecell{Inference Time \\ ($10^{-4}$ s/sample) ($\downarrow$)}  &\makecell{GPU Memory \\Usage (GB) ($\downarrow$)}\\
        \midrule
        \textit{LSTM} &-- &0.8332±0.0047

        &0.5840±0.0132 &-- &-- &\textbf{0.286}  &-- \\
        \midrule
        \multirow{4}*{\textit{ECSeq}} 
        &100 &0.8377±0.0023
        &0.6111±0.0077 &\textbf{5.318} &\textbf{9.950} &0.614 &\textbf{1.306}\\
        &500 
        &\textbf{0.8383±0.0018}
        &\textbf{0.6153±0.0079}
        &15.444 &9.955 &0.618  &1.664 \\
        &1,000 &0.8372±0.0021
        &0.6115±0.0083 &36.686 &10.480 &0.622 &2.283 \\
        &5,000 &0.8343±0.0013
        &0.6056±0.0055 &137.570 &28.930 &0.630 &7.037 \\
        \midrule
        \textit{batch GNN} &145,236 (1,000/batch) &0.8335±0.0017
        &0.5861±0.0157  &-- &10.675 &2.243 &6.414 \\
        \textit{full graph GNN} &145,236  &\multicolumn{5}{c}{Out-of-Memory}\\
        \bottomrule 
	\end{tabular}
 \end{adjustbox}
\end{table*}

\begin{table*}[htbp]
\caption{Performance of \textit{ECSeq} variants on \textit{FD2} and \textit{Bike}. \textit{R-Transformer} cannot converge on \textit{Bike}, so we do not report the results.
 } 
	\label{tab:variants}
	\centering
    \begin{adjustbox}{max width=\linewidth}
		\begin{tabular}{cccccccccccc}
		\toprule
        &\multirow{2}[2]{*}{\makecell{\textbf{Sequence Embedding}\\ \textbf{Extractor}}} &\multirow{2}[2]{*}{\makecell{\textbf{Graph Compression} \\ \textbf{Algorithm}}} &\multirow{2}[2]{*}{\textbf{GNN Model}} 
        &\multicolumn{2}{c}{\textbf{FD2}} & \multicolumn{2}{c}{\textbf{Bike}}\\ 
        \cmidrule(lr){5-6} \cmidrule(lr){7-8}
        & & & 
        & AUPRC ($\uparrow$) 
        & $\text{R@P}_{0.9}$ ($\uparrow$) &RMSE ($\downarrow$) 
        &sMAPE ($\downarrow$) 
        \\
        \midrule
        \textit{Sequence Model}&LSTM &-- &-- &0.7124±0.0076
        &0.6987±0.0078 &1.5463±0.0504 
        &0.1782±0.0040 
        \\
        \textit{w.o. Graphs}&R-Transformer 
        &--
        &-- &0.7064±0.0149
        &0.5403±0.1951
        &-- 
        &--  
        \\
        \midrule
        \multirow{8}{*}{\textit{ECSeq}} &LSTM &$k$-means &GraphSAGE (Mean)  
        &\textbf{0.7249±0.0112} &\textbf{0.7039±0.0032}
        &1.4832±0.0209 
        &0.1766±0.0038
        \\
        &R-Transformer &$k$-means &GraphSAGE (Mean)  &0.7105±0.0184
        &0.6991±0.0031
        &-- 
        &--
        \\
        &LSTM &AGC &GraphSAGE (Mean)  &0.7232±0.0102
        &0.6935±0.0095 &\textbf{1.4792±0.0218} 
        &0.1764±0.0044 
        \\
        &LSTM &Grain &GraphSAGE (Mean)  &0.7232±0.0102
        &0.6870±0.0095
        &1.4949±0.0219 
        &0.1812±0.0050
        \\
        &LSTM &RSA &GraphSAGE (Mean) &0.7201±0.0091
        &0.6922±0.0120
        &1.4812±0.0190 
        &\textbf{0.1761±0.0040}
        \\
        &LSTM &$k$-means &GraphSAGE (Max)  &0.7162±0.0083
        &0.7026±0.0049
        &1.4846±0.0166 
        &0.1768±0.0037 
        \\
        &LSTM &$k$-means &GCN  &0.7212±0.0102
        &0.6987±0.0052
        &1.9226±0.0376
        &0.2139±0.0059
        \\
        &LSTM &$k$-means &GAT  &0.7200±0.0112
        &0.7026±0.0026
        &2.0056±0.0107
        &0.2383±0.0053
        \\
        \bottomrule 
	\end{tabular}
 \end{adjustbox}
\vspace{-1.5em}
\end{table*}

Table \ref{tab:scalability} displays the AUPRC and R@P$_{0.9}$ results for fraud detection and the corresponding time consumption on \textit{FD1}. It is worth noting that training GNN on the full graph, which contains up to 145 thousand nodes, would result in an out-of-memory issue on our GPU with 11 GB RAM. However, by using graph compression, we can achieve satisfactory performance improvement ($\sim 5\%$ increase in R@P$_{0.9}$) while using only around 1GB of GPU RAM (100 compressed nodes).
We conduct a comparison by training a GNN model on a relation subgraph that includes nodes only in the batch, denoted as \textit{batch GNN} (the same strategy is used in \textit{GRASP}). The batch size is set to 1000, occupying approximately 6 GB of GPU RAM. The performance improvement of \textit{batch GNN} over \textit{LSTM} is insignificant and much worse than that of \textit{ECSeq}.


Importantly, Table \ref{tab:scalability} shows that the computational overhead of using \textit{ECSeq} beyond \textit{LSTM} is minimal. When compressing the original graph to 500 nodes (default setting) using $k$-means, the time consumption, including both compression and GNN training, is 25.4 seconds. If we further reduce the graph to 100 nodes, the time consumption is only 15.3 seconds. The inference time consumption is still at the $10^{-5}$ second-scale per sequence, similar to \textit{LSTM}. These results demonstrate the practicality of using \textit{ECSeq} in real systems.

Based on our observations, simply increasing the number of compressed nodes may not necessarily result in improved prediction accuracy. Therefore, we recommend that practitioners initially experiment with a smaller number of nodes, typically ranging from 100 to 500, as these tend to perform well across a variety of tasks.



\subsubsection{\textbf{Flexibility}}

We conduct experiments to evaluate the flexibility of \textit{ECSeq} framework in modifying sequence embedding extractors, graph compression methods, and GNN algorithms. The results are presented in Table~\ref{tab:variants}.

When either LSTM or R-Transformer is used as the sequence embedding extractor, applying \textit{ECSeq} can greatly enhance prediction performance. For example, in \textit{FD2}, the $\text{R@P}_{0.9}$ metric is increased from 0.6987/0.5403 to 0.7039/0.6991 for LSTM/R-Transformer respectively, by using \textit{ECSeq}. This confirms that the \textit{plug-and-play} characteristics of \textit{ECSeq} can flexibly enhance existing sequential modeling models that do not account for sequence relations.

Additionally, we study the performance of \textit{ECSeq} when using different variants of graph compression and GNN methods. Results show that different variants may have varying performance on different tasks. For example, using AGC as the graph compression method can further reduce prediction errors in the \textit{Bike} dataset as compared to the default $k$-means. The flexibility of \textit{ECSeq} allows us to easily change the compression and GNN algorithms, in case certain methods perform better than the default setting.

\begin{table*}[t]
 \caption{The new sequences of fraudulent behavior and their corresponding representative sequences with the closest relation (\textit{i.e.}, the highest cosine similarity of sequence embedding). Events are arranged from left to right in chronological order, and the underlined elements are the target events. \textit{RouterMac}: the mac address of users' devices; \textit{Item}: the category of goods; \textit{CardType}: the type of payment card (\textit{e.g.} credit card); \textit{Country}: the transaction location.} 
	\label{tab:interpretability}
	\centering
	\resizebox{\linewidth}{!}{
		\begin{tabular}{cc|ccccc|cccccccccc}
			\toprule
			&\textbf{Fields} & &\multicolumn{3}{c}{\textbf{New Sequences (Fraud)}}& & \multicolumn{5}{c}{\textbf{Representative Sequences with the Closest Relation}} \\ 
			\midrule
			\multirow{2}*{(a)} & \textbf{RouterMac} & MAC1 & MAC1 & MAC1 & MAC1 & \underline{MAC2} & MAC3 & MAC3 & MAC4 & MAC3 & \underline{MAC2} \\
			\cmidrule(lr){2-12}
			& \textbf{Amount} & 242 & 484 & 242 & 242 & \underline{5634} & 429 & 529 & 6 & 129 & \underline{5478} \\
			\midrule
			\multirow{2}*{(b)} & \textbf{Item} & I1 & I1 & I1 & I1 & \underline{I3} & I2 & I1 & I1	& I1 & \underline{I3}\\	
			\cmidrule(lr){2-12}
			& \textbf{Time} & 7:43 PM & 6:21 PM & 8:10 PM & 10:41 PM & \underline{3:05 AM} & 8:57 PM & 5:21 PM & 5:34 PM & 9:01 PM & \underline{2:06 AM} \\
			\midrule
			\multirow{2}*{(c)} & \textbf{Country} & C1 & C2 & C1 & C1 & \underline{C3} & C1 & C1 & C4 & C1 & \underline{C3}\\	
			\cmidrule(lr){2-12}	
			& \textbf{CardType} & TYPE1 & TYPE1 & TYPE1 & TYPE2 & \underline{TYPE3} & TYPE1 & TYPE2 & TYPE1 & TYPE1 & \underline{TYPE3} \\	
			\bottomrule 
	\end{tabular}}
 \vspace{-1.5em}
\end{table*}

\begin{figure*}[t]
 \centering	
   \subfloat[\textit{FD1}]{
   \includegraphics[width=0.36\linewidth]{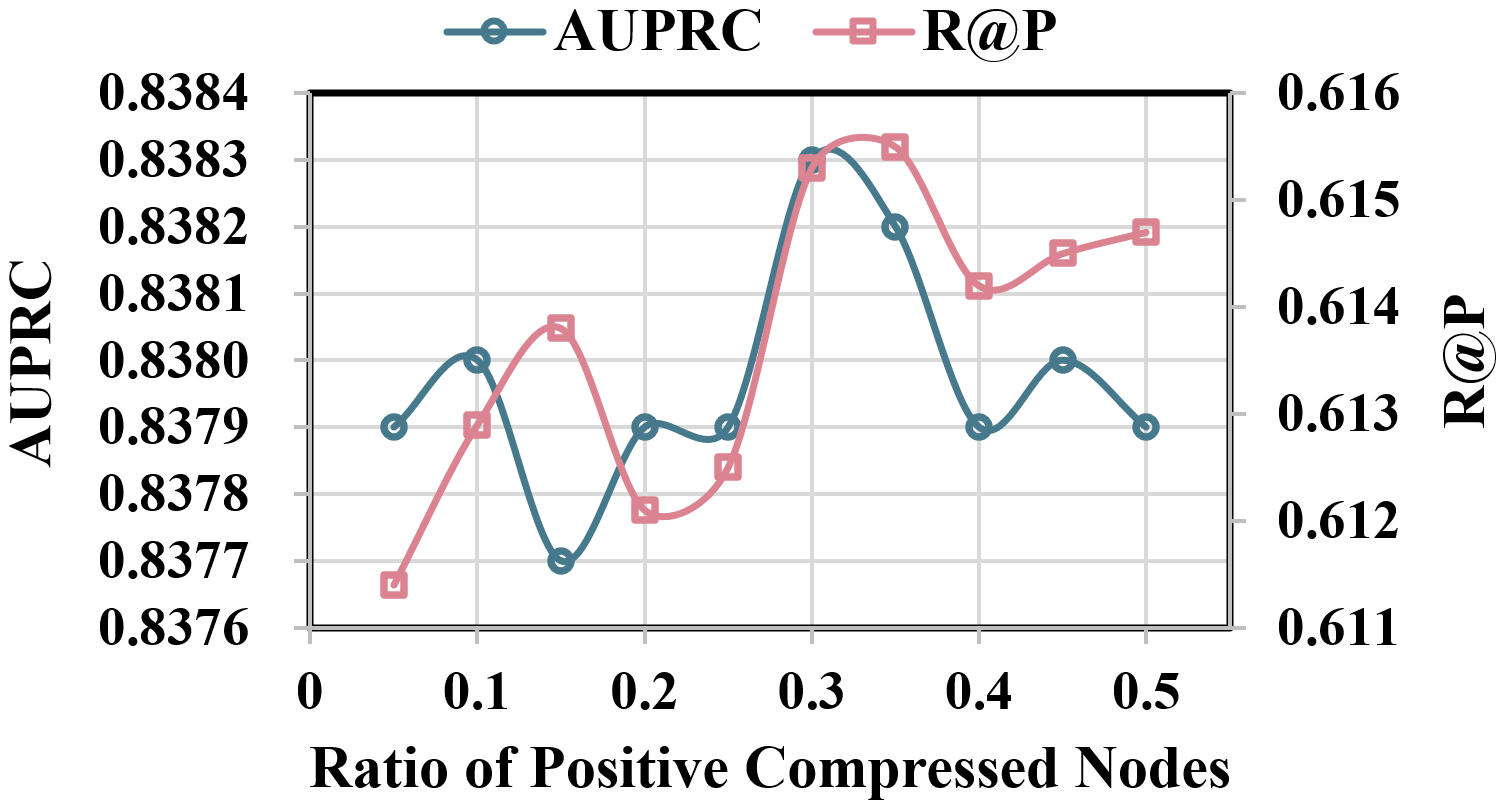}
   }
 \hspace{4em}
   \subfloat[\textit{FD2}]{
   \includegraphics[width=0.36\linewidth]{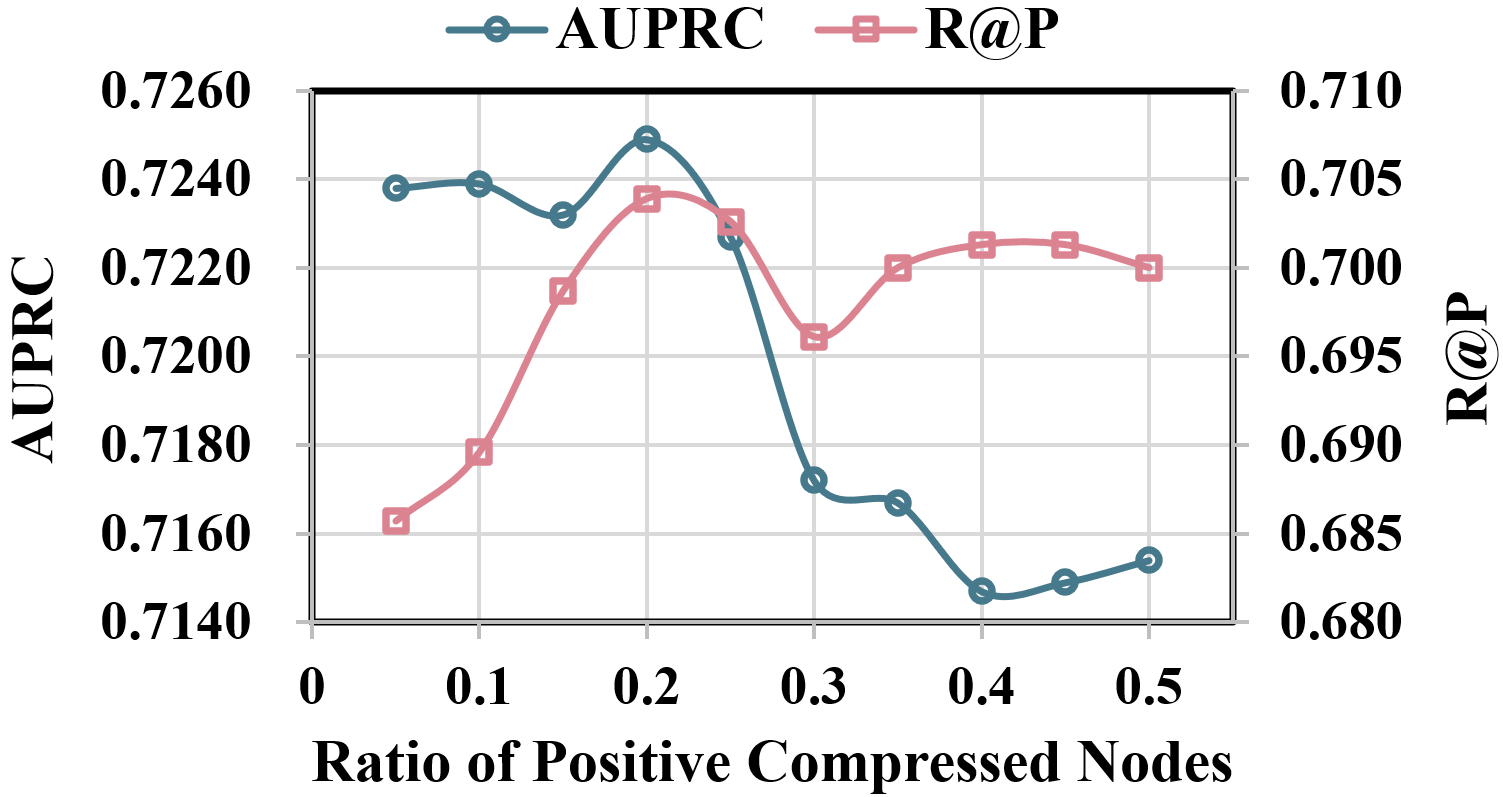}
   }
 \caption{Fraud detection performance of \textit{ECSeq} under varying settings of positive compressed node ratio.}
 \label{fig:sample_balancing}
 \vspace{-1em}
\end{figure*}

\subsubsection{\textbf{Sample Balancing}}


Figure~\ref{fig:sample_balancing} depicts the relationship between the ratio of positive compressed nodes and AUPRC/$\text{R@P}_{0.9}$ on \textit{FD1} and \textit{FD2}. As the ratio of positive samples escalates from the original ratio of 0.10/0.05 to a perfectly balanced ratio of 0.5, both AUPRC and $\text{R@P}_{0.9}$ first rise and then decrease. 
The optimal performance on \textit{FD1} manifests when the positive ratio approximates 0.3, almost tripling the original ratio of 0.1; while the zenith of AUPRC/$\text{R@P}_{0.9}$ on \textit{FD2} occurs when the positive ratio is around 0.2, roughly quadrupling the original ratio of 0.05.

It is believed that appropriately amplifying the ratio of positive compressed nodes can prioritize the knowledge gained from scarce but important positive samples, thus alleviating the sample imbalance problem and improving the performance of fraud detection.

\subsubsection{\textbf{Case-based Reasoning}} 

Here we analyze \textit{ECSeq}'s interpretability. 
First, we convert compressed nodes into representative nodes (sequences): (1) For coreset selection compression methods that select real nodes (\textit{e.g.}, Grain), the compressed nodes are real sequences, and we simply regard them as representative sequences. (2) For coreset selection methods that pick virtual synthetic nodes (\textit{e.g.}, $k$-means) or graph coarsening techniques, the compressed nodes lack corresponding existent sequences. Hence, we identify their source nodes based on the assignment matrix $\Phi$, and utilize the most similar real source nodes to approximate the compressed nodes. For instance, we may choose the source nodes with the highest attribute cosine similarity to the compressed nodes. Then we consider these selected real source nodes as representative sequences.

Subsequently, we construct connections between new sequences and the representative sequences based on the similarity of sequence embedding.
Table~\ref{tab:interpretability} manifests a set of new sequences of fraudulent behavior and their most strongly connected representative sequences (\textit{i.e.}, the representative sequences with the highest embedding cosine similarity). From these sequences, we have the following findings:

\begin{itemize} 
\item \textbf{Abnormal Amount}. From sequence (a), the transaction amount of the target event is larger than other events (5634 and 5478). Meanwhile, the target event occurs in the device `MAC2', which is different from the usually used device `MAC1'. It represents a situation: fraudsters log into the victim's account on a new device and attempt to transfer a large amount of money.  
\item \textbf{Abnormal Time}. From sequence (b), the transaction time of the target event is significantly different from other events. The target event of original sequences and similar sequences both occur at midnight (3:05 AM and 2:06 AM), which is contrary to people's daily routines. 
Note that the previous events all occur in daylight. It reveals a fraud pattern: 
fraudsters prefer to transfer money at midnight because people cannot respond to risk information in time.
\item \textbf{Abnormal Switching}. From sequence (c), we see that the payment card `TYPE3' used in the target event is different from the card in other events `TYPE1'. Moreover, the target event occurs in a different country, denoting the transaction request was sent from a remote abnormal location. It represents a situation: the fraudsters log in to a different country and try to transfer the money from a specific type of card.
\end{itemize}

The similar sequences exhibit common patterns, validating \textit{ECSeq}'s interpretability. We also discover three typical fraud patterns by analyzing the behavior characteristics of representative sequences, which may help us further develop explainable rules for fraud detection.

\section{Related Work}

As \textit{graph compression} techniques have been comprehensively discussed in Section \ref{sub:seq_relation_model}, here we review the related work in other two aspects, \textit{sequence representation learning} and \textit{graph neural networks}.

\subsection{Sequence Representation Learning}
A sequence refers to an ordered arrangement of elements, where the order of these elements often carries significant importance. 
To excavate sequence data, representation learning methods capture sequential dependencies within data, and transpose sequences into embedding vectors, which are utilized to execute prediction tasks, such as fraud detection~\cite{wang_2017_pkdd,jurgovsky_sequence_2018}, user mobility~\cite{chen2001use,do2019survey}, and item recommendation~\cite{Hidasi_session_based_2015,Beutel_latentcross_2018}.
Numerous effective models have been put forth to seize the sequential pattern and produce representation from users' historical information, including convolutional neural network (CNN)~\cite{fu_credit_2016,tang2018personalized}, recurrent neural network (RNN)~\cite{jurgovsky_sequence_2018,wang_2017_pkdd}, and self-attentive models~\cite{kang2018self,sun2019bert4rec,wang2019r}. 
To be specific, CLUE~\cite{wang_2017_pkdd} captures detailed information on users’ click behaviors, and models sequences of such clicks employing RNN to detect fraudulent transactions on e-commerce platforms.

Compared to these existing studies, our work tries to introduce graphs to explicitly model relations between behavior sequences. We note that some recent works (\textit{e.g.}, GRASP~\cite{Zhang_Gao_Ma_Wang_Wang_Tang_2021}) also model the relation graph between behavior sequences, but they do not address the efficiency and scalability issue when the relation graph becomes huge, thereby posing challenges to be deployed to online services.

\subsection{Graph Neural Networks} 
In recent years, graph neural networks (GNNs) have received widespread attention for successfully modeling complex patterns in graph-structured data. 
As representative GNN methods, GCN~\cite{kipf2016semi} propagates neighborhood information iteratively based on a first-order approximation of spectral convolutions;
GraphSAGE~\cite{HamiltonYL17} constitutes an inductive framework capable of generating node embedding for previously unseen data; GAT~\cite{velickovic2018graph} utilizes self-attention layers to compute discrepant weights for different neighbors during aggregation. 

GNN methods have shown success in multiple applications, including recommendation system~\cite{wang2019knowledge}, anomaly detection~\cite{jiangAnomaly2019,deng2021graph}, and traffic prediction~\cite{chen2020multi}.
In particular, current GNN-based approaches for learning user behavior sequences can be categorized into two main streams based on the construction of the graphs: some methods build sequence-level graphs by leveraging specific features of the sequences, such as the co-occurrence of sequence features~\cite{liang_sigir_2019} and the social relationships between users~\cite{zhong2020financial}; other methods captures fine-grained information by considering event-level interactions derived from event attributes~\cite{liuIntention,chang2021sequential}. 
Our work aligns with the first stream: we consider sequences as nodes, and the relation graph can be built in multiple ways, encompassing the universal approach of sequence embedding similarity and other domain-specific methods such as user friendships.

However, the time complexity of most existing GNNs is quadratically proportional to the number of nodes, or linearly related to the number of edges~\cite{wu2020comprehensive}, posing a challenge to directly utilize current GNN-based methods for mining massive sequence data.
While some techniques accelerate GNNs on large-scale graphs (\textit{e.g.}, Sketch-GNN~\cite{ding2022sketch} proposes a sketch-based algorithm whose training time and memory grow sublinearly to node count), our work tackles efficiency issues from an alternative angle -- graph compression, which also boasts the benefits of sample balancing and interpretability.

\section{Conclusion and Limitation}
Our proposed framework, named \textit{ECSeq}, aims to improve user behavior sequence learning by integrating sequence relations, while maintaining high efficiency and scalability for applicability in online services. The framework consists of two modules: \textit{sequence embedding extraction} and \textit{sequence relation modeling}, which enhances training and inference efficiency and provides a \textit{plug-and-play} capability. Specifically, with an extra training time of tens of seconds in total on 100,000+ sequences and inference time maintained within $10^{-4}$ seconds/sample, \textit{ECSeq} enhances the prediction R@P$_{0.9}$ of the widely used LSTM by $\sim 5\%$.

Currently, \textit{ECSeq} has a limitation where it can only deal with one type of relation. We aim to simultaneously incorporate more types of real-life relationships, such as users' social networks and behavioral habits similarity, into our framework in the future. We believe that the integration of different relationships can greatly enhance the relation modeling module. Our next step is thus to explore how to effectively and efficiently incorporate heterogeneous relationships.

\section*{Acknowledgements}
This research was supported by National Key R\&D Program of China (2023YFB3308504) and Ant Group.

\clearpage
\bibliographystyle{IEEEtran}
\bibliography{ecseq}

\end{document}